# FlowAR: une plateforme uniformisée pour la reconnaissance des activités humaines à partir de capteurs binaires

Ali Ncibi*,**, Luc Bouganim*,**
Philippe Pucheral *,**

*Univ. de Versailles/St-Q., 45 av. des Etats-Unis, 78035 Versailles cedex, France
**INRIA Saclay, 1 rue H. d'Estienne d'Orves, 91120 Palaiseau, France
<prenom.nom>@inria.fr

**Résumé.** Cette démonstration présente une plateforme expérimentale dédiée au développement de systèmes de reconnaissance des activités humaines (*Activity Recognition*), notamment celles de la vie quotidienne (AVQ), à partir de données issues de capteurs tels que les capteurs binaires. En adoptant une approche orientée données, cette plateforme nommée *FlowAR* s'appuie sur un pipeline (*flow*) de traitement des données structuré en trois étapes clés : nettoyage, segmentation et classification personnalisée. Grâce à sa modularité, elle offre une flexibilité pour explorer diverses méthodes et jeux de données, permettant des évaluations comparatives rigoureuses et une adaptation aux besoins spécifiques de développement. Un cas d'usage concret illustre son efficacité.

## 1 Introduction

Nous nous intéressons au développement orienté données d'un système de reconnaissance d'activités de la vie quotidienne (AVQ) à partir de capteurs binaires. Ce travail est réalisé dans le cadre du projet Domycile, associant l'équipe PETRUS, Hippocad (La Poste) et le Département des Yvelines. Il vise le suivi des patients dépendants, principalement des personnes âgées, via 10 000 box dotées de bases de données médico-sociales et enrichies par des capteurs. Les capteurs enregistreront les interactions des patients avec des objets ou zones domestiques stratégiques. Ces données, transmises via une connexion locale sécurisée à chaque box, seront analysées par le système embarqué pour détecter les AVQ et des changements propres à alerter les professionnels médico-sociaux (Voir Fig.1). Pour limiter l'intrusion, les données seront traitées au sein de chaque box grâce à un système sécurisé de gestion de base de données personnelles, pour n'externaliser que les tableaux de bord et alertes (Anciaux et al., 2019).

Le système de prise de décision envisagé repose sur un processus structuré en trois étapes fondamentales : le nettoyage des données capturées (élimination du bruit), leur segmentation, et la reconnaissance des activités par classification des segments. L'efficacité globale du système dépend fortement de la dernière étape, qui s'appuie sur des modèles d'apprentissage automatique capables de gérer les incertitudes liées aux capteurs et aux comportements observés. La personnalisation de ces modèles, essentielle pour répondre aux besoins spécifiques de chaque utilisateur, pose également un défi majeur et nécessite une préparation minutieuse des données d'entraînement.

FlowAR: Plateforme pour la reconnaissance d'activités humaines avec capteurs

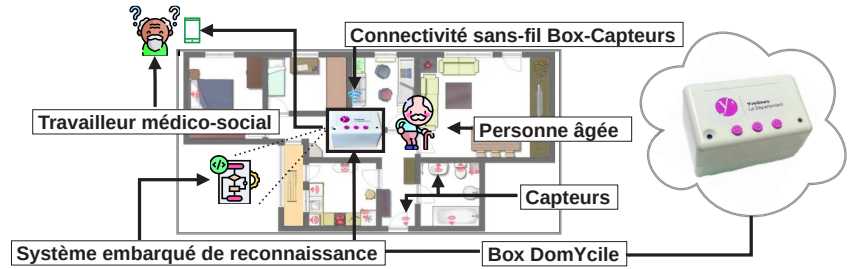

FIG. 1 – *Illustration du contexte du projet.*

Dans cette démarche, nous exploitons des ensembles de données publiques issus d'études similaires passées pour pallier l'impossibilité de collecter des données expérimentales auprès de patients réels. Cependant, ces données sont hétérogènes et requièrent un outil systématique pour les rendre comparables et exploitables dans notre contexte. Par ailleurs, malgré la diversité des travaux de recherche portant sur ce domaine, cette étude nous confronte à plusieurs défis : (1) *la reproductibilité* : en raison d'un manque d'outils pour répliquer et évaluer rigoureusement les résultats des expériences publiées ; (2) *l'explicabilité* : certains modèles utilisés pour la classification personnalisée sont souvent opaques, posant problème notamment dans les applications médico-sociales où les décisions doivent être transparentes et compréhensibles ; (3) *le manque de justification* : les choix techniques, notamment en segmentation des données et sélection de représentations pour la classification sont fréquemment présentés sans justification robuste ; et (4) *l'évaluation des performances* : les métriques classiques comme le F1-score ne sont pas toujours adaptées pour juger de l'efficacité des modèles dans certains contextes.

Pour répondre à ces défis, nous proposons une plateforme expérimentale qui uniformise le traitement des données et facilite l'exploration méthodique de nouvelles propositions mais aussi des solutions de l'état de l'art. L'intérêt de cette plateforme expérimentale dépasse le strict cadre de l'étude DomYcile et peut s'appliquer à de nombreuses études confrontées aux mêmes défis. Nous commençons par décrire les dimensions du processus décisionnel et les défis des approches actuelles, avant de présenter un pipeline conçu pour uniformiser les données et le développement orienté par celles-ci. Ce pipeline est intégré dans la plateforme opérationnelle *FlowAR*, illustrée à travers un scénario concret. Enfin, nous concluons en discutant des enseignements tirés et des perspectives d'amélioration.

## 2  Dimensions du problème

Le développement orienté données du système envisagé repose sur plusieurs étapes interdépendantes.

**Nettoyage.** Cette étape vise à identifier et corriger les anomalies potentielles des données capturées, telles que les activations intempestives ou prolongées. Des techniques de clustering, influencées par les représentations, mesures de similarité et seuils de regroupement (Iftikhar et al., 2020 ; Ye et al., 2016), aident à distinguer événements normaux et anomalies. Il est essentiel de différencier les anomalies à corriger de celles signalant des activités anormales nécessitant une alerte, qui ne doivent pas être supprimées.

**Segmentation.** Le flux d'observation continue doit être transformé en unités temporelles discrètes. La segmentation explicite identifie des transitions entre activités via détection de points de changement (Bermejo et al., 2021), tandis que la segmentation implicite utilise une fenêtre glissante centrée sur des fragments temporels, sans viser une correspondance précise avec les frontières d'activités (Krishnan et Cook, 2014). Quelle que soit l'approche, une segmentation de qualité nécessitera un ajustement itératif de paramètres (e.g., taille de la fenêtre glissante).

**Représentation.** Dès lors, les segments peuvent être décrits par des variables telles que la durée du contexte, les états et transitions observés, les fréquences d'activation, ainsi que les localisations et interactions des capteurs (Aminikhanghahi et Cook, 2019; Medina-Quero et al., 2018; Tapia et al., 2004; van Kasteren et al., 2008). En complément, des méthodes d'apprentissage profond enrichissent ces représentations, bien que moins interprétables et dépendantes des données d'entraînement (Hwang et al., 2021).

**Classification.** La classification des segments s'effectue enfin via des modèles multi-classes ou binaires, incluant souvent une classe pour l'inactivité ou les activités non spécifiées. La variabilité des durées et intensités peut entraîner un déséquilibre des classes, nécessitant des ajustements des données ou des modèles (Fahad et al., 2015). Le choix des modèles, ponctuels ou séquentiels, dépend des exigences de l'application (qualité et volume des données, précision, etc.). L'intérêt pour l'explicabilité croît, renforçant la transparence des décisions de certains modèles de classification opaques (Das et al., 2023).

**Personnalisation.** Afin de personnaliser ces modèles, une approche directe consiste à les entraîner avec des données annotées des utilisateurs cibles (Tapia et al., 2004; van Kasteren et al., 2008), en réduisant le coût d'annotation via des techniques comme l'apprentissage actif ou le clustering (Alemdar et al., 2017; Hiremath et al., 2022). Pour refléter l'évolution des activités individuelles dans le temps, les modèles doivent s'adapter afin de prévenir les dérives potentielles (Ordóñez et al., 2013; Abdallah et al., 2019). À défaut d'annotations personnelles suffisantes, des modèles initiaux basés sur des environnements similaires peuvent être affinés avec des données, même non annotées, de l'utilisateur (Cook, 2010; van Kasteren et al., 2010). Cela nécessite une abstraction manuelle des capteurs, bien que des techniques optimisées soient encore étudiées (Yu et al., 2022).

**Évaluation.** L'évaluation du système repose sur l'analyse des erreurs de classification, telles que les faux positifs et négatifs, souvent synthétisées par des scores classiques (e.g., Accuracy, F1-score, etc.). Pour prendre en compte la dimension temporelle, des erreurs spécifiques, comme l'insertion, la suppression et la fragmentation, sont explorées (Ward et al., 2011). Une estimation robuste des scores s'obtient généralement par validation croisée adaptée aux données temporelles, comme la méthode leave-one-day-out (Alemdar et al., 2015).

La diversité des options rend complexe la conception d'une solution complète, nécessitant des ajustements itératifs et des analyses approfondies pour optimiser les choix et performances.

## 3 Pipeline du traitement

Dans cette section, nous présentons une contribution clé de ce travail : un pipeline structuré qui uniformise les données publiques collectées, optimise leur exploitation et assure un processus de développement flexible et reproductible.

FlowAR: Plateforme pour la reconnaissance d'activités humaines avec capteurs

**Uniformisation des données.** Les données collectées, issues de travaux sur les AVQ et capteurs binaires, proviennent de foyers réels et sont annotées selon les activités des résidents (Alemdar et al., 2017; Cook, 2010; Ordóñez et al., 2013; Tapia et al., 2004; van Kasteren et al., 2008), mais sans format standardisé. Les capteurs utilisés sont des dispositifs ON/OFF discrets et peu intrusifs qui détectent des événements simples (e.g., ouverture de porte, pression de matelas, présence humaine). Dans notre travail, la structuration homogène choisie pour ces données inclut l'heure de début et de fin, un identifiant unique, un ordre chronologique des événements et la prise en compte des chevauchements pour refléter les situations réelles (Voir Fig.2.A). Les activités annotées couvrent des tâches quotidiennes (e.g., repas, sommeil, toilette) réalisées par un ou plusieurs résidents. La structuration des annotations pour un résident suit des principes similaires aux données capteurs, intégrant des stratégies pour gérer les chevauchements, parfois résolus par une règle simple de séparation facilitant la classification.

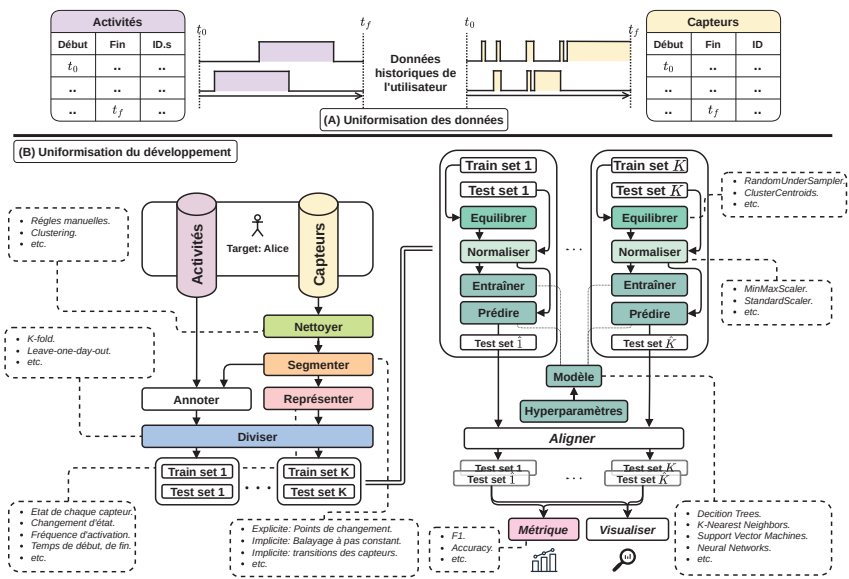

FIG. 2 – *Uniformisation du développement orienté par les données d'un système de reconnaissance d'activités humaines (e.g., AVQ) sur des traces de capteurs (e.g., binaires). Le mode de personnalisation repose directement sur des données annotées spécifiques à l'utilisateur.*

**Uniformisation du développement.** Comme le suggère la section 2, le développement orienté données des systèmes de reconnaissance d'activités humaines (e.g., AVQ) basées sur des capteurs (e.g., capteurs binaires) suit un pipeline structuré intégrant nettoyage, segmentation, représentation, classification personnalisée et évaluation. Dans notre projet, la personnalisation sera introduite progressivement. Nous analysons d'abord l'impact de choix de segmentation, représentation et classification sur les performances dans un scénario idéal (données annotées), puis adaptons ces choix aux scénarios sans annotations personnelles. La figure 2.B illustre ce processus supervisé, où des données personnelles annotées développent les composants, no-tamment le module de reconnaissance. Les capteurs subissent un nettoyage, les segments sont annotés selon les pré-annotations et représentés conformément aux choix retenus. Les modèles de classification sont ensuite entraînés et évalués via une validation croisée. Les performances sont interprétées pour identifier les forces et faiblesses du système. Ce processus, reproductible sur divers jeux de données simulant profils et environnements variés, évalue robustesse et fiabilité. Le pipeline permet des ajustements progressifs en modifiant paramètres ou méthodes (e.g., fenêtre de segmentation, représentations, modèles, techniques de validation croisée, métriques), assurant une analyse approfondie de l'impact de chaque configuration sur la qualité du développement.

## 4 Plateforme opérationnelle

Afin de concrétiser ce pipeline, une plateforme opérationnelle, accessible en ligne [1], a été développée avec deux composants principaux : une interface graphique (GUI) et un backend. La GUI offre l'accès à l'environnement et gère deux cas d'usage : (1) exploration des données harmonisées via des visualisations enrichies et statistiques descriptives (durées, fréquences, chevauchements) et (2) expérimentations de développement orientées données. Ces expérimentations suivent deux phases : *configuration et exécution* pour paramétrer et lancer les expériences dans le backend, et *évaluation* pour analyser les performances. Les résultats intermédiaires, accessibles en temps réel, sont sauvegardés pour une évaluation ultérieure. Des règles manuelles de nettoyage des données filtrent le bruit si nécessaire, permettant une comparaison avant/après nettoyage. Développée en Python, la plateforme utilise Streamlit pour la GUI, simplifiant la création d'interfaces web enrichies optimisées pour la manipulation et visualisation des données. Le backend repose sur Scikit-learn, PyTorch, Pandas pour le traitement et la modélisation, ainsi que Plotly et Matplotlib pour les visualisations. Sa modularité facilite l'ajout de données ou fonctionnalités à chaque étape du pipeline.

## 5 Démonstration

Après avoir fait un tour rapide des fonctionnalités offertes par la plateforme sur chaque étape du pipeline, nous développons un scénario d'usage illustratif. Pour faciliter la compréhension et l'interprétabilité des résultats, plusieurs choix sont simplifiés. Nous utilisons des données annotées sans anomalies apparentes. Une segmentation implicite est réalisée via une fenêtre glissante centrée sur les transitions de capteurs, chaque segment étant associé à l'activité dominante. Les segments sont décrits par l'état des capteurs (activé ou non) sur la période concernée. Pour la classification, un arbre de décision multi-classes est utilisé sans rééquilibrage des données, ce qui permet une analyse directe de la pertinence de la représentation choisie. Une validation croisée, basée sur une variante leave-one-day-out, entraîne le modèle uniquement sur les jours précédant un jour de test. Le score F1 est utilisé, à titre d'exemple, comme métrique d'évaluation. Cette configuration est intégrée à la plateforme avec d'autres options méthodologiques.

La plateforme propose une sélection de jeux de données harmonisés (13 à ce jour) et permet d'en intégrer de nouveaux au format uniformisé (voir Fig. 2.A). Pour notre scénario, nous utilisons la fonctionnalité d'exploration, incluant une vérification visuelle de l'absence de bruit via des graphiques synthétiques illustrant les durées d'activation des capteurs. Nous sélectionnons alors deux ensembles pertinents : Ordonez-A (13 jours) et Ordonez-B (21 jours), issus du même projet mais capturés dans deux foyers distincts avec capteurs et activités similaires en nature et nombre. En appliquant la configuration définie, nous obtenons un score F1 moyen de 79% pour Ordonez-A. Une analyse des matrices de confusion et des erreurs temporelles (fragmentation, fusion, décalage) générées automatiquement corrobore ce résultat. Cependant, sur Ordonez-B, le score F1 diminue à 49%, avec des confusions importantes entre activités. Grâce aux outils d'analyse de la plateforme (graphique des chevauchements capteurs-activités et distribution des valeurs de représentation selon les segments), nous identifions une cause probable : les capteurs de porte, souvent activés lors des transitions entre pièces, introduisent des ambiguïtés dans la représentation. Pour tester cette hypothèse, nous désactivons ces capteurs via la plateforme et relançons l'expérimentation. Les performances s'améliorent significativement à 59%. Ce scénario révèle une limitation généralisable : les représentations fondées sur l'état des capteurs perdent en efficacité lorsqu'elles incluent des capteurs liés à des transitions ou événements non spécifiques aux activités d'intérêt. Ce constat, déjà souligné dans des études antérieures, peut être atténué en enrichissant les représentations avec des encodages des transitions d'état (van Kasteren et al., 2008). Une vidéo de démonstration illustre le scénario, mettant en avant la flexibilité de la plateforme pour analyser les choix de données, identifier les limitations de performances, et ajuster les configurations en fonction des spécificités des jeux de données, de manière rapide et efficace.

---

1. `https://flowar.alincibi.me`

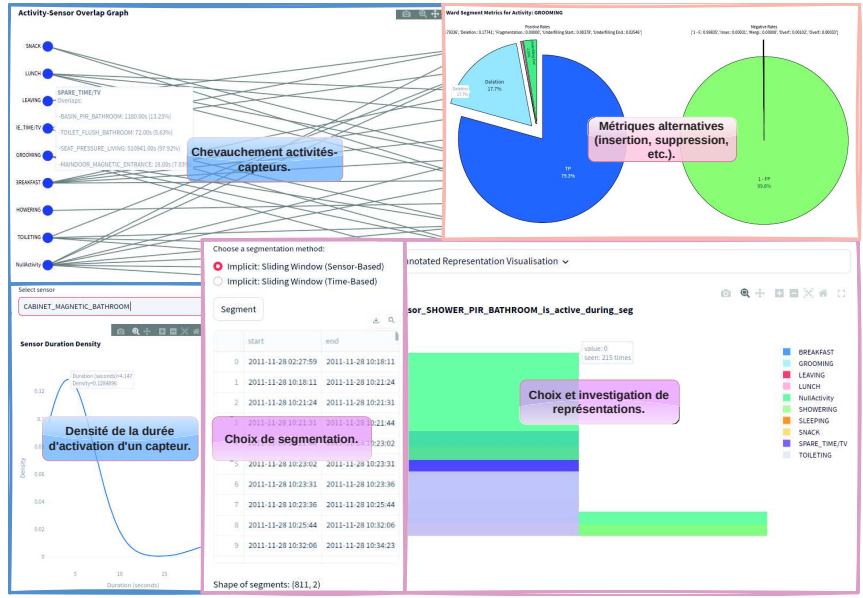

FIG. 3 – *Exemples de captures d'écran de la plateforme.*

# 6 Conclusion

Notre plateforme propose une approche structurée pour le développement de systèmes de reconnaissance d'activités humaines à partir de données de capteurs. Bien qu'illustrée pour les AVQ et les capteurs binaires, elle demeure applicable à d'autres types de capteurs et d'activités. Elle offre un cadre flexible permettant un cycle itératif de configuration, d'expérimentation et d'analyse, standardisant et accélérant ainsi le processus de développement. À court terme, elle permettra de calibrer le système sélectionné dans le projet Domycile, renforçant ainsi son rôle dans l'amélioration du suivi médico-social des personnes dépendantes. L'évolution vers des scénarios non annotés permettra d'intégrer des techniques avancées de personnalisation dans le pipeline, élargissant encore la flexibilité et les applications de la plateforme. Ces avancées contribueront à maintenir un développement rigoureux, reproductible et adaptable à des contextes diversifiés.

## Remerciements

## Références

## Summary

This demo showcases a platform for developing human activity recognition (AR) systems, focusing on daily activities using sensor data, like binary sensors. With a data-driven approach, this platform, named *FlowAR*, features a three-step pipeline (flow): data cleaning, segmentation, and personalized classification. Its modularity allows flexibility to test methods, datasets, and ensure rigorous evaluations. A concrete use case demonstrates its effectiveness.